%% file: main.tex
\documentclass{article}

\usepackage{arxiv}
\usepackage[utf8]{inputenc}
\usepackage{natbib}
\setcitestyle{numbers,open={[},close={]}} %Citation-related commands
\usepackage{xcolor}
\usepackage{bm}
\usepackage{graphicx}
\usepackage{hyperref}
\usepackage{xspace}
\usepackage{ctable} % tables
\usepackage{colortbl} % for line coloring in tables
\usepackage{makecell}
\usepackage{acronym}
\input{acronyms}

\title{Explainable Machine Learning with Prior Knowledge: An Overview}

\author{
Katharina Beckh\thanks{Equal contribution}~\thanks{Corresponding author: katharina.beckh@iais.fraunhofer.de} \\
Fraunhofer IAIS \\
\And
Sebastian Müller\footnotemark[1] \\
University of Bonn
\And
Matthias Jakobs\footnotemark[1] \\
TU Dortmund University \\
\And
Vanessa Toborek\footnotemark[1] \\
University of Bonn
\And
Hanxiao Tan\footnotemark[1] \\
TU Dortmund University \\
\And
Raphael Fischer\footnotemark[1] \\
TU Dortmund University \\
\And
Pascal Welke \\
University of Bonn
\And
Sebastian Houben \\
Fraunhofer IAIS \\
\And
Laura von Rueden \\
Fraunhofer IAIS \\
}

\begin{document}

\maketitle

\begin{abstract}
This survey presents an overview of integrating prior knowledge into machine learning systems in order to improve explainability. The complexity of machine learning models has elicited research to make them more explainable. However, most explainability methods cannot provide insight beyond the given data, requiring additional information about the context. We propose to harness prior knowledge to improve upon the explanation capabilities of machine learning models. In this paper, we present a categorization of current research into three main categories which either integrate knowledge into the machine learning pipeline, into the explainability method or derive knowledge from explanations. To classify the papers, we build upon the existing taxonomy of informed machine learning and extend it from the perspective of explainability. We conclude with open challenges and research directions.
\end{abstract}

\input{ch_1_introduction}

\input{ch_2_background}

\input{ch_informed_exp_intro}
\input{ch_informed_exp_main1}
\input{ch_informed_exp_main2}

\input{ch_informed_exp_main3}

\input{ch_x_discussion}

\paragraph{Acknowledgements}
This research has been funded by the Federal Ministry of Education and Research of Germany as part of the Competence Center Machine Learning Rhine-Ruhr ML2R (01|S18038ABC).

\bibliography{literature_clean}

\end{document}

%% file: acronyms.tex
\newacro{ml}[ML]{machine learning}
\newcommand{\ml}{\ac{ml}\@\xspace}

\newacro{iml}[IML]{informed machine learning}
\newcommand{\iml}{\ac{iml}\@\xspace}

\newacro{ai}[AI]{artificial intelligence}
\newcommand{\ai}{\ac{ai}\@\xspace}

\newacro{cnn}[CNN]{convolutional neural network}
\newcommand{\cnn}{\ac{cnn}\@\xspace}

\newacro{am}[AM]{activation maximization}
\newcommand{\am}{\ac{am}\@\xspace}

%% file: ch_1_introduction.tex
\section{Introduction}
\label{sec:intro}

The complexity of current \ml models makes the precise decision process during inference difficult to grasp for humans.
Most models can only be scrutinized in the correlation of input and output features, leaving their internal workings opaque.
This is why they are often referred to as black-box models.
Particularly in high-stake scenarios, the resulting lack of interpretability poses a severe drawback.
For example, consider applications in which an AI-supported system predicts patient sepsis, or where a system rejects a loan. In these scenarios, insight into the decision process is important to ensure safety, fairness and compliance with legislation \cite{regulation2016regulation}.
As a consequence, recent work focuses on explainability to improve transparency and trustworthiness of machine learning models~\cite{adadi2018peeking, Li2020, arrieta2020explainable, Burkart2021}.
While explanation capabilities have been investigated since the development of expert systems \cite{confalonieri2021historical}, the advancements in research and the prevalent use of \ml systems lead to new requirements and expectations regarding explanations.
The computational steps a model takes may exactly describe \textit{how} the algorithm comes up with a prediction, but the typical questions are more concerned with the \textit{why}, asking for causal or contrastive explanations \cite{Miller2019}.

Explanations are by nature context-sensitive as there is no explanation without a question and no question without context.
The context of an explanation encompasses not only the matter that is to be explained but also the recipient of the explanation \cite{ras2018explanation, norkute2021, chari2020directions}. Thus, following from the range of scenarios that require explanations, a diverse group of recipients needs to be considered.
Explainability methods do not only have to bridge a communication gap between ML systems and experts~\cite{doran2017does}, but their outputs also need to be accessible to a diverse group of users and meet their respective requirements. The context-sensitive nature of the task makes it inherently difficult to develop a generalized explainability method that can be automatically deployed under all circumstances.

If the explainability method itself is not context-aware, fulfillment of this necessity is delegated to the user: \citet{Molnar2020a} point out pitfalls of explainability methods that require additional knowledge by the user in order to gain reliable new insights and to prevent false conclusions.
These methods are part of a larger class of data-driven explainability methods that produce \textit{feature attribution} values for a prediction. Feature attribution refers to the effect and importance of data features on the model prediction. 
These methods can detect a \textit{correlation} but they do not provide an answer on \textit{why} a feature is relevant to the model output. To answer this question the user has to apply their own knowledge to put the results into context.
However, a layperson might not have the necessary knowledge or the feature values might be intrinsically difficult to interpret, e.g., in the case of raw sensor data like audio signals.
Furthermore, these explanations are data-constrained implying that they cannot provide insight beyond the data at hand~\cite{Roscher2019}.\footnote{To underline the argument of data-constraints, Zachary Lipton states that desiderata (e.g. fairness, explainability) concerning real-world tasks are not captured in the prediction sandbox, https://www.youtube.com/watch?v=fvL6MSzsQ6Q\&t=1522s, Accessed 16.04.2021.}

We suggest the integration of prior knowledge to overcome those limitations and to provide the user with the necessary context.
This knowledge is often expressed in a comprehensible form, e.g., as knowledge graphs or logical rules.
The integration of prior knowledge was already motivated in the 2000s in connection with support vector machines \cite{lauer2008incorporating} and has been revisited recently for scientific discoveries from data and \ml output \cite{karpatne2017theory,Roscher2019}.
Recent research introduced the notion of \iml, which offers a comprehensive taxonomy on the integration of prior knowledge into \ml \cite{VonRueden}.
In their work, the authors focused on the classical accuracy-driven learning pipeline and identified explainability as a possible side effect of knowledge integration.
Independently, \citet{Li2020} have covered this informed perspective and provide a distinction between data-driven and knowledge-aware explainable \ml.
They subdivide the knowledge-aware approaches into broad categories of general knowledge methods and knowledge-based models.
The structure from \citet{Li2020} mainly considers approaches from a method-centric standpoint while possibilities for knowledge integration are not discussed.
A systematic overview on how the prior knowledge is actually integrated to benefit explainability does not exist yet.

In this paper, we present approaches that harness prior knowledge to make machine learning models more explainable and therefore usable.
To this end, we use an existing taxonomy on informed machine learning \cite{VonRueden} to structure relevant literature.
Our contributions are summarized as follows.
\begin{itemize}
    \item We present three ways of how the integration of prior knowledge benefits explainability in existing work.
    \item We provide an overview of different knowledge integration approaches to facilitate the application and adaptation of these methods.
    \item We highlight open challenges and research directions.
\end{itemize}

\begin{figure}[!t]
    \centering
    \includegraphics[width=.85\textwidth]{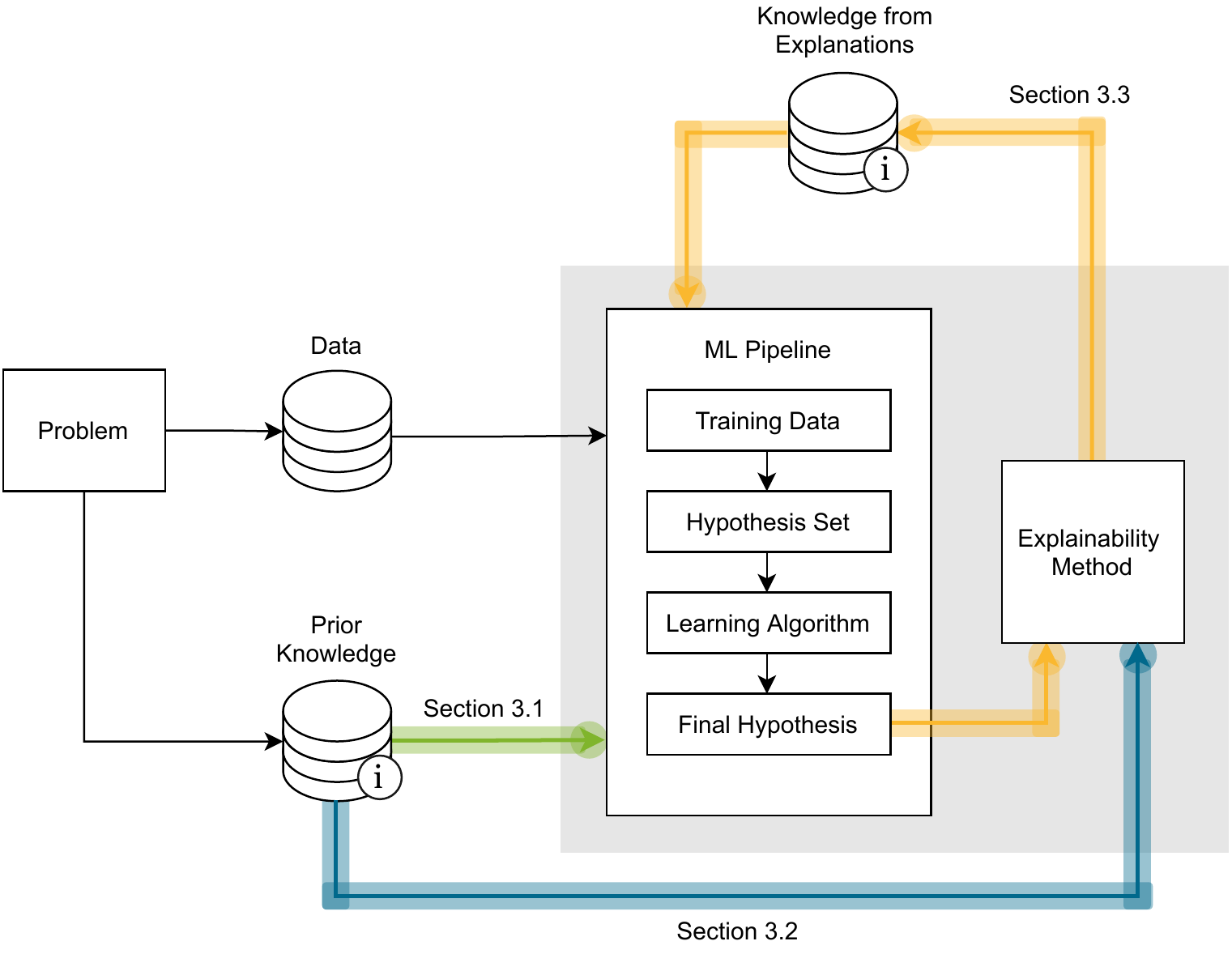}
    \caption{Framework of explainable machine learning from an informed ML perspective. Three different ways of integrating prior information (depicted by databases with information symbols) are discussed in the corresponding sections of this paper.}
    \label{fig:framework}
\end{figure}

The framework of knowledge-driven explainable machine learning is schematically displayed in \autoref{fig:framework}, and also outlines the core structure of our paper.
The figure shows three ways to integrate knowledge. The first approach, shown in green, is to integrate knowledge into the machine learning pipeline. Blue exemplifies the integration of knowledge into the explainability method. The yellow arrow shows how knowledge can be derived from explanations and then be integrated into the machine learning pipeline.
In \autoref{sec:background} we define and provide background on explainability and introduce informed machine learning with its respective taxonomy.
In \autoref{sec:main} we investigate approaches on how prior knowledge can inform explainable machine learning.
A discussion is provided in \autoref{sec:discussion} followed by open challenges and research directions.
Finally, we summarize our findings in \autoref{sec:conclusion}.

%% file: ch_2_background.tex
\section{Background}
\label{sec:background}

Next, we give an overview of current methods to make complex machine learning models more understandable by providing explanations.
In addition, we highlight some limitations of these methods, mainly the fact that measuring their quality is a fundamentally hard endeavor. Afterwards, we describe a recent formalism to incorporate prior knowledge into the machine learning pipeline called \iml \cite{VonRueden} which builds the basis for our work.

\subsection{Explainable Machine Learning}

Explanations constitute an important part of human interactions because, in a societal context, humans are interested in the motivations behind a decision~\cite{Lipton2018a}.
Following the work by \citet{Miller2019}, an explanation describes the process of abductive inference as well as the final product, i.e., the answer to a why-question.
With the increasing application of machine learning models, explanations are required for multiple reasons: verification of the system, improvement of the system, learning from the system and compliance to legislation~\cite{Samek2017XAI, regulation2016regulation}.

We base our definition of explainable machine learning on prior work \cite{Rudin2019,Kim2018a} in which the authors draw a distinction between \textit{interpretable} and \textit{explainable} machine learning models. The former describes models that demonstrate an inherent transparency, while the latter describes models that are incomprehensible by themselves but gain transparency through explanations created by methods dedicated to understanding how the model works~\cite{Burkart2021, Rudin2019}.
Note that the degree to which a model is inherently interpretable is strongly dependent on the model size and choice of input features. A small decision tree is self-explanatory and thus constitutes an interpretable model. However, with increasing model size, the tree becomes less interpretable, forcing the user to resort to explainability methods in order to obtain insights into the model behavior.
Moreover, if the input features to a model do not correspond to human semantic concepts, the interpretability of the decision process suffers as a result, no matter how simple the model is. Therefore, we agree to differentiate the interpretability of the model at three different levels,  either at the level of the entire model, individual components or the training algorithm according to~\cite{Lipton2018a}, but also highlight the integration of additional knowledge as one interpretable component of the model (cf. \autoref{subsec:IML}).

Currently favored models such as neural networks and random forests exemplify the need for explainability methods since even small versions are too complex to be interpretable in the above defined sense.
In order to explain the behavior of a  black-box model for individual predictions, there are three main approaches:
First, attribution-based methods were developed to estimate the respective contribution of each of the input features to the prediction~\cite{Sundararajan2017,Lundberg2017,BinderAlexanderMontavon2016,Ribeiro2016}.
In the area of computer vision these values are often visualized as heatmaps.
Some of these methods are theoretically grounded in Game Theory via the concept of Shapley values~\cite{Shapley1953,Aumann1974}, which quantify the contribution each player has on an outcome~\cite{Lundberg2017,Sundararajan2017}.
One downside of these approaches is that they merely show on which part of the input the model placed importance but not for which reasons~\cite{Stammer2020}.
The second approach consists of training an interpretable surrogate model, e.g., a decision tree or a linear model, to mimic the black-box as closely as possible, including its errors~\cite{Frosst2017,Tan2018}.
As mentioned before, the interpretability of the surrogate model is itself dependent on its size.
If a big surrogate model is needed to accurately mimic the complex model behavior it will not be useful in practice.
The third approach comes from the recent push towards counterfactual explanations, following literature from social science, as pointed out by ~\citet{Miller2019}.
Miller argues that humans tend to ask \textit{"What if ...?"} counterfactual questions to gain an understanding of the underlying decision process behind an external rationale.
Counterfactual explanations rely on this theory to construct artificial data points, which are close to the data point to explain but are classified differently~\cite{Wachter2017,Dandl2020}.
In this manner they provide the user with a minimal set of changed features needed for a different outcome.
One of the main challenges is the generation of plausible yet minimally changed counterfactual examples~\cite{Dandl2020}.

While all these approaches aim to explain the unknown inner workings of a complex model, it is often difficult to judge whether or not they are accurately following the model's line of decision or not.
Ground truth explainability data cannot be provided in most scenarios since knowledge about the decision process of the model is needed, which is exactly what explainability methods try to uncover.
Some work looked into presenting users with explanations generated from explainability methods and observed how well they help users make the initial prediction \cite{hase2020-evaluating, buccinca2020}.
However, it is unclear whether or not an explanation that appears to make a model more understandable to the user is also the correct explanation.
One way to remedy this conundrum of missing ground truth data might be to incorporate prior knowledge into the training and explanation process.
By providing the model with information about how to navigate the path from input to prediction, it might learn a human-understandable way and become more interpretable.
\par

\subsection{Informed Machine Learning}
The motivations for integrating prior knowledge into the machine learning pipeline can be manifold. A natural goal is to improve the model performance or to train with less data. With trustworthy \ai becoming more important~\cite{brundage2020toward}, another purpose of \iml is to ensure knowledge conformity or to improve the interpretability of a model~\cite{VonRueden}.

In informed learning, prior knowledge is integrated into the learning process.
In contrast to traditional \ml that uses domain knowledge, e.g., for feature engineering, informed machine learning makes the integration of prior knowledge more explicit in order to improve the otherwise mainly data-driven approaches.
\iml can be defined as learning from a hybrid information source that consists of data and prior knowledge, in which the prior knowledge stems from a data-independent source, is given by formal representations, and is explicitly integrated into the machine learning pipeline~\cite{VonRueden}.

The taxonomy of \iml provides a framework for classifying its different approaches with respect to the knowledge source, the knowledge representation, and the integration stage in the learning pipeline~\cite{VonRueden}. According to the authors, the spectrum of informed learning is described in terms of the following building blocks:
The \textit{source} can be either scientific knowledge (like natural sciences or engineering), world knowledge (like vision, linguistics, or general knowledge) or intuitive expert knowledge.
The knowledge can be formalized using \textit{representations} such as equations, simulations, rules, or graphs, but it can also be given more informally via human feedback. These representations can be \textit{integrated} into one of the four stages of the learning pipeline (cf. \autoref{fig:framework}):
\begin{itemize}
    \item \textbf{Training data}: In contrast to the typical way of incorporating knowledge via feature engineering, an informed approach is defined as \textit{hybrid} by using both the original data set and an additional, separate knowledge source.
    \item \textbf{Hypothesis set}: The integration into the hypothesis set is accomplished through the selection of architecture and hyperparameter settings.
    \item \textbf{Learning algorithm}: Through a loss function, i.e., an appropriate regularizer, additional knowledge can be integrated into the learning algorithm.
    \item \textbf{Final hypothesis}: Existing knowledge can be used to compare, benchmark and post-process the output of a model.
\end{itemize}

In the described \iml taxonomy interpretability is considered a side effect. Extending the information flow gives rise to directions of informed learning for explainable methods.

%% file: ch_informed_exp_intro.tex
\section{Knowledge-driven Explainable Machine Learning}
\label{sec:main}

As motivated in the introduction, explanations always have to bridge a communication gap between the model and the receiver of the explanation. If the explanation fails to communicate in concepts intelligible to the receiver, the receiver has no chance to understand the explanation. Furthermore, we have noted that purely data-driven explanations put the responsibility to draw reliable conclusions fully on the user, thereby limiting the relevant user group to experts.
We propose to address both shortcomings by integrating prior knowledge
into the \ml pipeline, or the explanation, to adapt the explanation to users and contexts.
Based on our literature search we identified three main approaches to incorporate prior knowledge for improving explainable learning systems (colored arrows in \autoref{fig:framework}):
\begin{enumerate}
    \item Informed Machine Learning to increase Explainability
    \item Informed Explainability
    \item Deriving Knowledge from Explanations
\end{enumerate}
The first one (\autoref{fig:framework} green arrow, \autoref{subsec:IML}) is well-captured by the \iml taxonomy.
It covers cases where some form of prior information is available in addition to data, and which is integrated at some stage of the learning pipeline.
In the works we review in this section, the integrated knowledge is used to align model components with domain knowledge. This facilitates model-handling for professionals from the domain, who might not even be \ml experts.
In the second case (\autoref{fig:framework} blue arrow, \autoref{subsec:IX}), knowledge integration takes place at the explainability method.
We consider this an additional component to the usual ML pipeline, due to the prevalence of post-hoc approaches. This integration type offers the largest potential to adapt explanations to user groups and contexts because it allows for interactions between users and model explanations.
Lastly, many approaches were found to derive additional information from explanation results (\autoref{fig:framework} yellow arrow, \autoref{subsec:derive}).
Similar to prior knowledge that is available from the start, these newly derived priors can then be re-integrated at any point of the pipeline. Approaches in this category are adapted by \ml scientists and developers to debug and improve models.
With the latter two strands, we go beyond the information flow established in the IML taxonomy \cite{VonRueden} and thus propose an extended framework, as shown in \autoref{fig:framework}.

We now discuss the related work that has been published for all strands.
During our investigations we found that many approaches can be neatly categorized via their knowledge integration type, i.e., training data, hypothesis set, learning algorithm, and final hypothesis.
Where applicable, we structure the subsections accordingly.

%% file: ch_informed_exp_main1.tex
\subsection{Informed Machine Learning to increase Explainability}
\label{subsec:IML}

\begin{figure}[h]
    \centering
    \includegraphics[width=.5\textwidth]{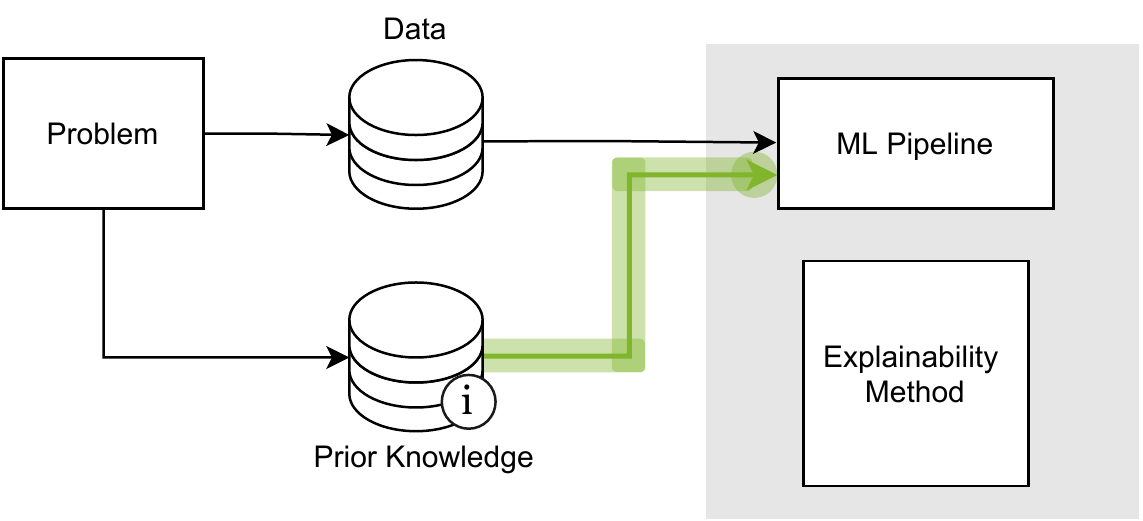}
    \caption{In addition to the data set used in the \ml setting, prior knowledge is integrated into the \ml pipeline (green arrow).}
    \label{fig:3.1-framework}
\end{figure}

In this category, additional information in the sense of \iml is used in such a way that it not only increases model performance but also improves explainability. While this improvement is not always stated as an explicit objective, we argue that often additional knowledge is integrated in the sense of an individual, interpretable component, therefore increasing the interpretability of the entire machine learning pipeline.
The remainder of this section is following the four stages of the learning pipeline as introduced earlier.

\paragraph{Training Data} We found two applications in the field of recommender systems that increase the interpretability by integrating additional, human-understandable knowledge into the training data for their models. The general task is to predict the next item(s) a user interacts with given an (ordered) set of user-item interactions. \citet{Wang2019d} and \citet{Ma2019} both first create a single heterogeneous knowledge graph by connecting the user-item interaction data with multiple existing knowledge bases (MovieLens-1M + IMDb and Freebase + DBPedia respectively). \citet{Wang2019d} generate recommendations by extracting limited-length user-item paths from the graph and rating them for plausibility using a recurrent neural network. \citet{Ma2019} compute sets of rules, where a rule is a sequence of certain types of edges, and learn a weighting of those rules. Both methods are argued to be explainable because the algorithms are forced to reason along the edges of the knowledge graph and produce a weighting that reflects the contribution of each path or rule to the decision.

\paragraph{Hypothesis Set}
We found several applications that promote the interpretability of the model by using domain knowledge to perform architectural changes in the \ml model. Two of them are located in the field of biology. \citet{Ma2019a} encode biological knowledge into the network in the form of factor graphs, representing either genes or gene ontology as neurons and abstracting direct influence to corresponding genes or gene ontologies as edges, giving semantic meaning to all originally meaningless neurons and their connections. Such a network, constructed based on prior knowledge rather than heuristics, is easily intelligible and therefore more explainable. To learn representations of single-cell RNA-sequence data, \citet{Rybakov2020} propose an interpretable autoencoder based on a regularized linear decoder. The autoencoder decomposes variations into interpretable components using prior knowledge in the form of annotated feature sets obtained from public databases. Observed covariates, such as batch or cell type, can be fed into the encoder-decoder architecture or simply weighted by a linear model and then introduced into the autoencoder. As the primary purpose of the method is to explain the components of variations, introducing prior knowledge enables more acceptable interpretations.

Prior information in the form of world knowledge improves \ml models for the problems of semantic image understanding \citep{Chen2012} and conversation generation \citep{Liu2020}. \citet{Chen2012} create a pipeline processing the visual cues of an image input as well as the background knowledge of a guide ontology. The result is a directed graphical model that constructs possible relationships between the visible objects. This \ml model becomes more interpretable because all resulting relationships are verifiable through the ontology.
\citet{Liu2020} aim to make the process of conversation generation more transparent by integrating a factoid knowledge graph into the deep learning pipeline that is augmented with information from related text documents. While the knowledge graph provides background knowledge for an encoder-decoder model, it also makes the \ml model more interpretable, because all graph traversals for knowledge selection can be retraced.

\citet{Chen2020} propose a replacement for batch normalization layers, commonly found in neural network architectures, called \emph{concept whitening} layers.
The datapoints flowing into the layer first get decorrelated using a whitening operation and then aligned in the latent space to a fixed number of predefined concepts.
As an example, the authors train a \cnn for image classification.
After swapping all batch normalization layers with concept whitening layers, they use a separate dataset, labeled with concepts such as \emph{aeroplane} and \emph{person}, to finetune the alignment of the training data to these concepts.
This alignment to human-interpretable concepts helps not only in debugging the training process, i.e., discovering missalignment between similar concepts, but also increases the understandability of the decision process, because it allows to break down the decision to a mixture of known and understandable concepts.

\paragraph{Learning Algorithm}
Several applications use insights from domain knowledge to apply regularizers in the learning algorithm.
The knowledge graphs used by these applications are medical ontologies such as ICD10 or SNOMED-CT that have a simple tree structure. \citet{Choi2017} and a subsequent extension proposed by \citet{Ma2018c} predict future diseases of a patient given the diagnosis history of that patient. They compute a vector embedding of the ontology and derive a feature representation of the inputs by accumulating relevant parts of the ontology embedding via an attention mechanism which can be seen as an explanation capability.
\citet{Jiang2019} use logistic regression to infer the readmission probability  of a patient after a hospital stay from their medical history. A distance measure over the ontology is included as a regularization factor in the loss function that penalizes biases towards a certain part of the ontology. \citet{Yan2019} formulate a joint learning task of multi-label assignment to CT images and retrieving images similar to the input from a database. The retrieved images serve as explanations. Mutually exclusive label combinations are extracted from the ontology and this information is used to regularize the loss function of the retrieval task, this way aligning the explanations closer to the ontology.

Another two approaches in the domain of computer vision integrate additional knowledge as constraints in the \ml pipeline. \citet{Donadello2017} present Logical Tensor Networks, a neural network for semantic image interpretation constrained by first-order fuzzy logic. These logic rules are derived from the comprehensible WordNet ontology describing part-of relations and are used to exclude classifications showing unrealistic relations like, e.g., \textit{tail} as a part of \textit{table}.
For the problem of part localization, \citet{Zhang2017} present an approach that leverages human feedback to improve a \ml model. The model consists of an And-Or graph that is based on a pre-trained \cnn. This graph disentangles the hierarchical relationship between semantic image parts (top level nodes) and single activated \cnn units (terminal nodes). In a second step, they visualize the network's activations using up-convolutional networks and evaluate them via human feedback. This additional information improves the And-Or graph by excluding activations that do not contribute to the target semantic part.

\paragraph{Final Hypothesis}
The following applications use external, human-understandable knowledge to perform plausibility checks on the results of the \ml models.
\citet{doran2017does} make a conceptual proposition to extend an already existing, explainable model by a post-processing step that checks the explanation against a knowledge base for plausibility. A rudimentary realization of this idea is to improve the output of a multi-object detector~\citet{Pommellet2019}. First, the authors define a number of categories the objects in the knowledge base can fall into. Subsequently, they compute a ranking that describes how closely related two categories are. To refine the output of the multi-object detector, they proceed to collect information about which categories are detected and artificially increase the certainty score for objects that belong to related categories.

\citet{Kim2018a} propose to utilize an external data set of predefined concepts, for example in the form of images, to test how much the networks latent representations align with these concepts. To do so, they take the latent representation of these concepts together with representations of random samples and train a linear classifier to distinguish between concept-related and random vectors. Using the normal to the decision boundary, they can quantify how much a data point from the original data set, used to train the model, aligns with the concepts. This approach is similar to \citet{Chen2020}, in that it improves the interpretability of the model by measuring the alignment of datapoints to human-understandable concepts in the model. However, whereas \citet{Chen2020} use the external concept dataset to finetune the network to ensure the alignment, \citet{Kim2018a} merely measure the alignment to the concepts after training the model in the conventional way.

\paragraph{Section Summary}
The reviewed papers show that the integration of knowledge can improve explanation capabilities.
For each stage in the pipeline the explainability can be improved by enforcing alignment to human understandable concepts.
For the training data, the data itself gets connected to semantic knowledge, e.g., in the form of knowledge graphs.
In the stages of hypothesis set and learning algorithm, the model structure or representations are aligned with prior knowledge which brings the internals of a model closer to comprehensible concepts.
In the final hypothesis stage, the prior knowledge enables a plausibility check for model explanations.

%% file: ch_informed_exp_main2.tex
\subsection{Informed Explainability}
\label{subsec:IX}

\begin{figure}[h]
    \centering
    \includegraphics[width=.5\textwidth]{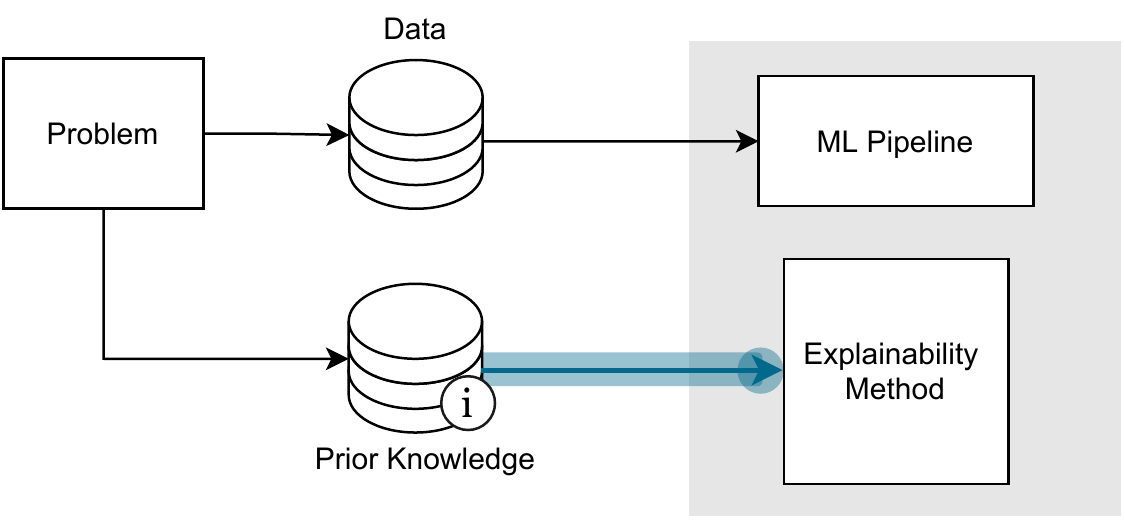}
    \caption{Prior knowledge is integrated into the explainability method (blue arrow).}
    \label{fig:3.2-framework}
\end{figure}

The research work presented so far integrates knowledge into the learning pipeline as presented in the \iml taxonomy \cite{VonRueden}.
Moving onwards, we now investigate cases in which prior knowledge helps designing and executing explainability mechanisms, which we purposefully do not consider part of the pipeline (cf. \autoref{fig:framework}).
This is an important and unique way of integrating knowledge into ML that extends the \iml taxonomy by adding the \emph{explainability method} as a new integration type.
For this,
we identified two main opportunities: formalized priors for explanations and interactive explanations.

\paragraph{Formalized Priors for Explanations}
\citet{shams2021rem} propose a methodology which extracts conditional rules from deep neural networks and combines them with other data-driven and knowledge-driven ones. Clinicians are able to directly validate and calibrate the extracted rules with their domain knowledge to yield more precise and acceptable explanations.

Generating counterfactuals is one direction of providing example-based explanations~\cite{stepin2021cf}.
While many approaches search counterfactuals for the original instances based on distance metrics, \citet{Mothilal2020} narrow down the search space via user-defined causality constraints.
They formalize their priors in the form of box constraints on feasible ranges.
Another work~\cite{Mahajan2019} attempts to assure the causal plausibility of counterfactuals by incorporating a penalty in the optimization process for infeasible values. For causal relationships that cannot be expressed with formulas, they train a variational autoencoder that generates counterfactuals and evaluates their quality based on human feedback. Both approaches promote the causal plausibility of generated counterfactuals by introducing prior knowledge into optimization in the form of algebraic constraints.

Another post-hoc explainability method is \am. It attempts to discover the ideal input distribution of a given class (global explanation) by optimizing the gradients of the inputs while freezing all parameters of the networks. However, \am with no priors tends to generate meaningless mosaics in high frequency, which are not human-recognizable. Two approaches~\cite{Yosinski2015,WeiZTF15} aim at improving the visual interpretability by adding $\alpha$-norm regularization to the optimizer. \citet{Mahendran2015} constrain the total variation of the explanation to prior images, thus, providing a smoother output. Moreover, \citet{Yosinski2015} penalize nonsensical high-frequency pixels by applying Gaussian blur kernels to activations before each optimization step. All of the above methods generate more human-understandable explanations by incorporating additional algebraic restrictions in the optimizers.

Another approach combines LIME \cite{Ribeiro2016} with Inductive Logic Programming \cite{muggleton1994} to obtain verbal explanations for image classification \cite{rabold2019enriching}. Extracting symbolic rules from images enables a different perceptual modality and more expressive explanations, such as spatial relations of image parts.

\paragraph{Interactive Explanations}
The \iml taxonomy considers human feedback as a valid representation type, with corresponding sources usually being world or expert knowledge.
This is rooted in the success of incorporating human interaction within learning processes.
As an example, the framework of coactive learning \cite{Shivaswamy2015} allows users to correct and thus improve model predictions via direct feedback.
Work has also been done on making the human interaction robust against manipulation, for example in the medical domain \cite{10.1007/978-3-319-23344-4_36}.
Other approaches use interactive visual analytics to analyze and possibly obtain information on how to refine trained models \cite{liu2017better,Paiva2015}.

Recently, human interaction was identified to possibly benefit the explainability of ML systems \cite{Sokol_2020}.
This is mainly driven forward by the observation that in communication between humans, personalized explanations or even explanation dialogues result in better understanding and acceptance \cite{Miller2019}.
Accordingly, enabling users to interact with provided explanations can potentially fulfill several explainability desiderata, i.e., requirements for explaining methods as identified by fact sheets \cite{Sokol2019}.

Different works have successfully developed ML systems that provide interactive explanations for their decisions.
The \emph{What-If Tool} \cite{8807255} allows for interactively analyzing a model and finding explanations via feature importance.
\citet{Krause2016} also allow for interactively exploring model decisions via underlying feature importance.
Moreover, they offer the option of tweaking feature values in order to see how it affects the predictions.
Customization of explanations was also successfully implemented in the \emph{MUSE} (Model Understanding through Subspace Explanations) framework \cite{Lakkaraju2019}.
It allows users to interactively choose features of interest, and then explore how the model behaves in resulting subspaces.
Another interactive explanation system is \emph{Glass-Box} \cite{Sokol2017}, which offers explanation dialogues via a voice-enabled virtual assistant.
\citet{schneider2019personalized} proposed a conceptualization for methods that shall provide personalized explanations.

Besides showing the mere feasibility of interactive explainable ML, those works also evaluated the impact on users.
They seemed to obtain a better and also faster understanding of the underlying model logic \cite{Lakkaraju2019}.
By using interactive tools, users were able to improve the predictive model quality \cite{Krause2016}.
Generally, users were satisfied with the received interactive explanations, but they criticized the lack of arguability \cite{Sokol_2020}.

%% file: ch_informed_exp_main3.tex
\subsection{Deriving Knowledge from Explanations}
\label{subsec:derive}

\begin{figure}[h]
    \centering
    \includegraphics[width=.65\textwidth]{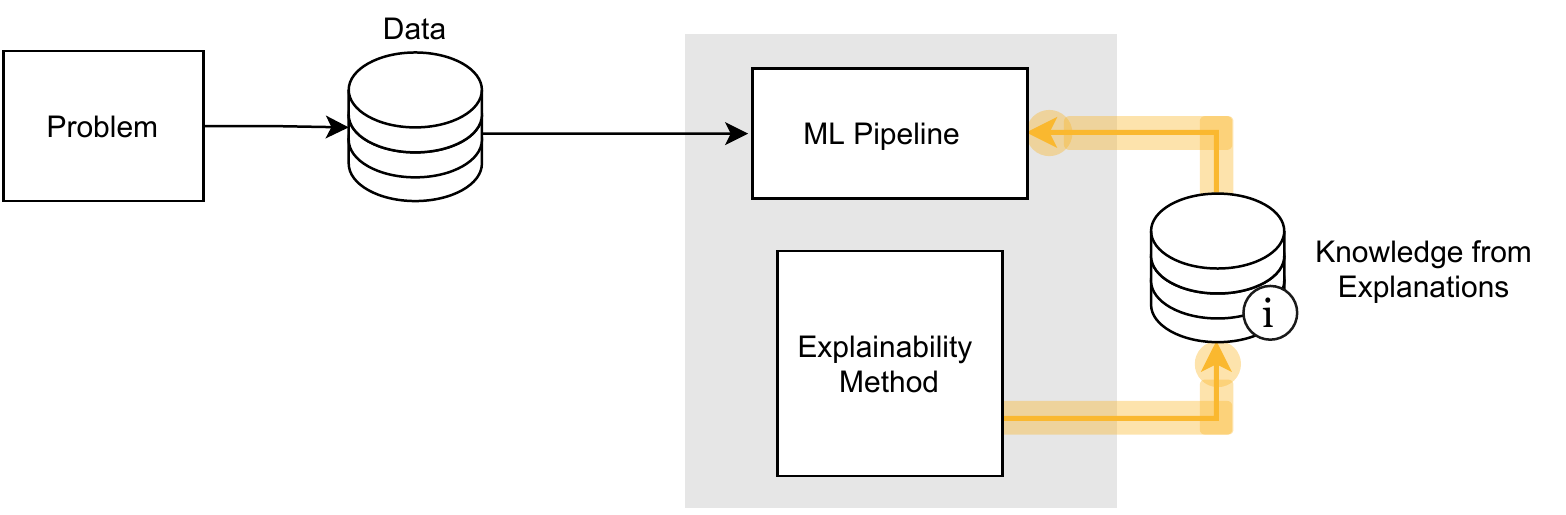}
    \caption{From the explanations of an explainability model, knowledge is derived, formalized and subsequently incorporated in the ML pipeline (yellow arrow).}
    \label{fig:3.3-framework}
\end{figure}

Explainability methods often detect flaws in \ml models, and as such, inform about necessary improvements.
One prominent use case is to remedy the Clever Hans phenomenon \cite{lapuschkin_unmasking_2019} which refers to some models latching onto spurious correlations in the data set, instead a known, correct relation that is present in the data and obvious to humans. As an example, the authors show an image containing a photographer's watermark in the bottom left corner on most images featuring a horse. The model picked up on that artifact and focused its decision upon the watermark being present instead of the horse.

We have found that many approaches suggest explicit ways of formalizing and incorporating feedback on explanations that go beyond simply showing the necessity of model improvement.
In essence, knowledge is derived from the explainability component of the model and is subsequently integrated into the learning pipeline (cf. \autoref{fig:3.3-framework}).
Related to this, a curated list gives an overview of supervision on model explanations\footnote{https://github.com/stefanoteso/awesome-explanatory-supervision}.
We present the papers that we found structured according to the knowledge integration types.

\paragraph{Training Data}
Some approaches use feedback on explanations for revising and improving the corpus of available training data in order to inform the learning system.
This is usually achieved with a human-in-the-loop, who inspects visual explanations for data instances, revises them if necessary, and thus adds more training data \cite{Baur2020}.
In terms of \iml, this can be understood as obtaining prior information from an expert user (or from world knowledge) at the final hypothesis which is represented as human interaction.
This information is then used to improve the training data for the next learning iteration.

\citet{Schramowski2020a} and \citet{Teso2019} generate additional training data using feedback on explanations.
If an expert decides that the explanation of the model is incorrect, the respective part of the image is used to create counterexamples. % expert changes explanation
Counterexamples are generated by randomizing or otherwise altering the parts an expert identified as unimportant.
Each counterexample, however, retains the label of the original datapoint, encouraging the model to unlearn the unwanted correlation.
Instead of adding new data points, the so-called \emph{Explanatory Debugging}~\cite{Kulesza2015} lets the user correct mislabelled data points in an interactive way during training.

\paragraph{Learning Algorithm}
The feature-based feedback from the Explanatory Debugging approach~\cite{Kulesza2015} is incorporated as a Bayesian prior, meaning that human feedback at the final hypothesis is transformed into a probabilistic representation, which is then integrated into the learning algorithm.

We found a line of research focusing on the integration of explanations in the learning algorithm by adding a regularizing term to a model's loss function.
In addition to the training loss between the model's prediction and the ground truth label, an additional loss between the model's explanation for each prediction and a given explanation is added and thus simultaneously minimized.

To the best of our knowledge, this approach was popularized by \citet{Ross2017}, who gather binary annotation masks from experts indicating parts of the input the model should not focus on.
These annotations are then used to penalize the gradient of the prediction w.r.t. to specific inputs which are nonzero in the annotation mask.
This forces the model to minimize the gradients at the selected locations as part of the training algorithm, which should then lead to the model being \emph{right for the right reason}~\cite{Ross2017,Schramowski2020a}.

\citet{rieger20a} use a decomposition-based approach to measure the importance of certain inputs to a model's decision. Using expert annotations similar to \citet{Ross2017} they force certain inputs to be regularized to have zero impact on the model.
They test their approach on skin cancer screening images by enforcing that colorful patches next to potentially cancerous skin lesions should be ignored since they do not inform about the type of lesion and only occur in one of the classes.
In evaluating their method, they found that the model significantly outperformed the approach of \citet{Ross2017} after being forced to ignore the colorful patches.
A similar approach was proposed by \citet{Selvaraju2019}, in which the authors evaluate the effect of human intervention on visual explanations in the application domains of visual question answering and image captioning.
They found, after incorporating the changed explanations, that their model was not only able to correctly highlight the images responsible for the correct answer or image caption but also outperform other state-of-the-art question answering systems.
In addition to visual feedback, \citet{Stammer2020} offer the user to provide semantic feedback in the form of relational functions.
Similar to \citet{rieger20a}, \citet{Erion2019} use attribution priors to optimize for desired explanation qualities, such as smoothness and sparsity, using the attribution method \textit{expected gradients}.
With this method, assumptions, e.g., that neighboring pixels should have a similar effect on the output, can be integrated into the model.

\citet{Balayan2020} provide a neural network-based framework that jointly makes predictions as well as associated explanations. Subsequently, the output is validated by human experts and the model is improved by adjusting its parameters through backpropagation. Using ground truth data the authors claim that human feedback increases the prediction quality of the explanations by over 13\%.
For a natural language inference task, \citet{Camburu2018} incorporate explanations of textual entailment, i.e., whether a premise sentence entails a hypothesis sentence,  into the training process using a negative log-likelihood for classification and an explanation loss.
Furthermore, a feature attribution method for text classification adds a loss with the goal to mitigate unintended bias in the text \cite{Liu2019a}. % L2 loss

To summarize, most approaches we found use a modified loss function, which takes into account feedback on the model's explanations, to regularize the model's behavior towards the desired outcome.

\definecolor{ca}{RGB}{166,97,26}
\definecolor{cb}{RGB}{223,194,125}
\definecolor{cc}{RGB}{128,205,193}
\definecolor{cd}{RGB}{1,133,113}

% ML2R colors
\definecolor{ml2rteal}{RGB}{0, 147, 145} % source: Powerpoint ML2R template
\definecolor{ml2rblue}{RGB}{1, 105, 140} % source: Powerpoint ML2R template
\definecolor{ml2rorange}{RGB}{250, 184, 47} % source: Powerpoint ML2R template
\definecolor{ml2rgreen}{RGB}{128, 181, 44} % source: Powerpoint ML2R template
\definecolor{ml2rdarkblue}{RGB}{0, 67, 112} % source: Poster template
\definecolor{greyline}{RGB}{180, 180, 180}

\colorlet{threeonecolor}{ml2rgreen!60}
\colorlet{threetwocolor}{ml2rblue!60}
\colorlet{threethreecolor}{ml2rorange!60}

\newcommand{\pillnc}[1]{\colorbox{gray}{#1}}
\newcommand{\threeonepillnc}[1]{\colorbox{threeonecolor}{#1}}
\newcommand{\threetwopillnc}[1]{\colorbox{threetwocolor}{#1}}
\newcommand{\threethreepillnc}[1]{\colorbox{threethreecolor}{#1}}

\newcommand{\pill}[1]{\pillnc{\footnotesize\cite{#1}}}
\newcommand{\threeonepill}[1]{\threeonepillnc{\footnotesize\cite{#1}}}
\newcommand{\threetwopill}[1]{\threetwopillnc{\footnotesize\cite{#1}}}
\newcommand{\threethreepill}[1]{\threethreepillnc{\footnotesize\cite{#1}}}

\renewcommand{\arraystretch}{1.8}
\begin{table*}[!b]
\begin{tabular}{lccccc}
\toprule
      & \makecell{Training\\ Data} & \makecell{Hypothesis\\ Set} & \makecell{Learning\\ Algorithm} & \makecell{Final\\ Hypothesis}  & \makecell{%(Post-hoc)\\
     Explainability\\ Method}\\
     \midrule
     Knowledge Graphs &  \threeonepill{Wang2019d} \threeonepill{Ma2019} & \threeonepill{Ma2019a} \threeonepill{Chen2012} \threeonepill{Liu2020} & \makecell{ \threeonepill{Choi2017} \threeonepill{Ma2018c} \threeonepill{Jiang2019} \\ \threeonepill{Yan2019} \threeonepill{Zhang2017}  \threethreepill{Erion2019} } &  \threeonepill{doran2017does} \threeonepill{Pommellet2019} & \\ \arrayrulecolor{greyline}\hline
     Logic Rules & & &  \threeonepill{Donadello2017} & & \threetwopill{rabold2019enriching}  \\ \arrayrulecolor{greyline}\hline
     Algebraic Equations & & \threeonepill{Rybakov2020} & \threethreepill{Erion2019} \threethreepill{rieger20a}&  \threeonepill{Kim2018a} &  \threetwopill{Mothilal2020} \threetwopill{Mahajan2019} \\ \arrayrulecolor{greyline}\hline
     Probabilistic Relations & & & \threethreepill{Erion2019} & & \\ \arrayrulecolor{greyline}\hline
     Human Feedback & \makecell{ \threethreepill{Baur2020} \threethreepill{Schramowski2020a} \threethreepill{Teso2019} \\ \threethreepill{Camburu2018} \threethreepill{Kulesza2015}} & & \makecell{\threethreepill{Ross2017} \threethreepill{rieger20a} \threethreepill{Selvaraju2019}\\ \threethreepill{Camburu2018} \threethreepill{Liu2019a} \threethreepill{Kulesza2015}} & \threethreepill{Balayan2020} & \makecell{ \threetwopill{shams2021rem} \threetwopill{Sokol_2020} \threetwopill{Krause2016}\\ \threetwopill{8807255}  \threetwopill{Lakkaraju2019} \threetwopill{Sokol2017} \\ \threetwopill{schneider2019personalized} } \\
     \arrayrulecolor{black}\bottomrule
\end{tabular}
\caption{Forms of knowledge integration as presented in \citet{VonRueden}; colors indicate how explainability is addressed or taken into account: \threeonepillnc{ \iml is used to increase explainability (\autoref{subsec:IML})}, \threetwopillnc{ knowledge is used to enhance the explainability method (\autoref{subsec:IX})}, \\ or \threethreepillnc{ explainability is used to derive and integrate knowledge (\autoref{subsec:derive})}.
}
\label{tab:overview}
\end{table*}

%% file: ch_x_discussion.tex
\section{Discussion and Outlook}
\label{sec:discussion}

\autoref{tab:overview} gives an overview of all papers presented in \autoref{sec:main} and advances their categorization according to the \iml taxonomy by considering the knowledge representation type. We want to point out that the "informed explainable" methods in \autoref{subsec:IX} constitute a new integration stage with respect to the \iml taxonomy. This results in an additional column solely populated by publications from that section.
The table shows a concentration of work using additional knowledge represented either as knowledge graphs or as human feedback. We did not find any work using simulation results or differential equations. Also, only two papers we identified deploy logic rules as representation type.

We now assess the strengths and limitations of integrating prior knowledge for explainability, both from a general perspective and for each of the three presented strands separately.

\paragraph{General Considerations}
%%%%%%%%%%%%%%%% summary %%%%%%%%%%%%%%%%
We established three strands for integrating prior knowledge for explainable machine learning. They can be distinguished with regards to how prior knowledge is available and at which point it is integrated. Prior knowledge is either given as independent data or it is derived from an explainability method. In the first case, the knowledge is then integrated into either the machine learning pipeline or the explainability method. In the latter, the knowledge is only integrated into the machine learning pipeline.

We hypothesize that the prevalence of knowledge graphs and human feedback as representation types is due to the intelligibility of both representation types. Knowledge graphs allow to represent scientific as well as world knowledge while their benefit lies in the structured representation of world knowledge. Human feedback, on the other hand, is the most accessible knowledge representation for expert knowledge, a knowledge type that is usually more intuitive and less formal. While this reasoning seems plausible to us, we cannot exclude the possibility of falling victim to a selection bias here.
Furthermore, the low number of papers using logical expressions was surprising since logic rules are inherently expressive\cite{muggleton2018}.
We found that many neuro-symbolic approaches are not informed in a sense that they first use an existing model and then try to improve it using additional knowledge. Instead, they aim to provide a concept that combines logic and connectionism (for an overview of that field we refer to \cite{Bouraoui2019}).

%%%%%%%%%%%%%%%% strengths %%%%%%%%%%%%%%%%
The key strength of integrating prior knowledge for explainability is that it improves the human understanding of the machine learning model.
Knowledge that is already available can be utilized to give context, address user needs and explanations can be used to inform the learning system.

Recall that some reasons for why we need explainable machine learning are verification of the system, compliance to legislation, improvement of the system, and learning from the system \cite{Samek2017XAI}. In the first two cases, verification of the system and compliance to legislation, an auditor assesses the behavior of the system. The auditor has to make an informed decision whether the system complies to all requirements without necessarily being an expert on the underlying technology.
Prior knowledge could serve as a bridge to equip the auditor with the required context to probe the system in a meaningful manner and come to a sound verdict.
Regarding the improvement of \ml systems, current explanation methods already help developers to detect flaws like the Clever Hans phenomenon.
Now, with the need for explanation, task performance is not the only objective that needs to be considered when designing a new system.
Although secondary objectives like fairness or transparency are elusive concepts
\cite{Krishnan2020a}, integrating prior knowledge could be a way to approach this problem (c.f.~\cite{Sokol_2020}).
Consider this: we do not train our models with fairness or explainability in mind, yet we fault them for not demonstrating these traits inherently. If we want to ensure fairness and explainability we need to clearly capture these in the learning process.

%%%%%%%%%%%%%%%% limitations %%%%%%%%%%%%%%%%
A prerequisite for informed machine learning is the availability of prior knowledge.
While not all domains have easy access to prior knowledge, a knowledge source that is not necessarily domain-specific may still be eligible for explainability. It is important to create awareness of existing knowledge bases and also to consider the different ways a knowledge source can be useful. For the latter point our work serves as a stepping stone for options.

%%%%%%%%%%%%%%%% future work %%%%%%%%%%%%%%%%
The current biggest challenge of explainability lies in its evaluation, and in this case to quantify how much a model improves when integrating prior knowledge. Since there is no formalism, the distinction we provide is rather broad with three levels of interpretability: inherently interpretable, interpretable components, and black-boxes.
Further research is needed to establish a more precise definition to be able to capture the different effects of incorporating prior knowledge on explainability.

The absence of methods that integrate differential equations and simulations
offers another starting point for future work. Both representation types are close to the field of physics, where informed machine learning is well-established \cite{raissi2019physics}, but the effect on explainability is less investigated. A possible direction is given by \citet{bikmukhametov2020combining} who incorporate first principles models and investigate the effect on the explainability.

\paragraph{Informed Machine Learning to Increase Explainability.}
%%%%%%%%%%%%%%%% summary %%%%%%%%%%%%%%%%%
In \autoref{subsec:IML}, prior knowledge is explicitly integrated into the machine learning pipeline to also improve model explainability.
While we have reviewed research in \autoref{subsec:IML} that is exemplary for each integration type, we can see that most approaches use the learning algorithm stage to integrate additional information.
Possibly, because the integration as regularization enables using existing model architectures and is less labor-intensive.

%%%%%%%%%%%%%%%% key strengths %%%%%%%%%%%%%%%%
The benefit of this approach is that it leverages existing knowledge and makes machine learning models more comprehensible by integrating the knowledge through an interpretable component into the machine learning pipeline. \citet{Choi2017} demonstrate a correlation between introducing the prior knowledge source and achieving more concise visualizations of embeddings as compared to competitors.

%%%%%%%%%%%%%%%% limitations %%%%%%%%%%%%%%%%
Again, the way in which the explanation quality is measured is often not addressed or evaluated.
From the papers that evaluated explainability in some regard there was no clear consensus on what type of measure is preferable.
We did not find a common understanding of the degree to which a method improves interpretability or explainability. In this sense, it is not straightforward to determine whether certain knowledge representations or integration types are especially effective. Reasons for this are the dependency on the application context and the lack of formalism of interpretability.

%%%%%%%%%%%%%%%% future work %%%%%%%%%%%%%%%%
There are many papers that can be categorized into \iml
which do not necessarily state it as an explicit goal to make their models more explainable.
To make machine learning models ready for use, explainability needs not be treated as an afterthought but has to already be considered in the design phase.

\paragraph{Informed Explainability.}
%%%%%%%%%%%%%%%% summary %%%%%%%%%%%%%%%%%
In \autoref{subsec:IX} prior knowledge is incorporated in the explainability method. We expanded on the informed machine learning taxonomy by introducing the explainability method as a new knowledge integration stage. A method qualifies for this category if it integrates an independent knowledge source in addition to an existing algorithm providing explanations.

%%%%%%%%%%%%%%%% key strengths %%%%%%%%%%%%%%%%
A distinction can be made between interactive approaches and formalized priors.
The benefit of the interactive approach is that the user can give direct feedback on an explanation.
This case can be seen as a communication module between the system and the user.
For the formalized priors, knowledge is integrated once to improve the explainability component directly.
In both cases incorporation of additional knowledge into the explainability method allows for accommodating user needs.

%%%%%%%%%%%%%%%% limitations %%%%%%%%%%%%%%%%
We emphasize that post-hoc methods should be considered with caution. Since explanations are obtained by approximations, it cannot be ensured that the explanations are faithful to the model. The integration of a comprehensible knowledge source does not change that and should not lead to a false sense of security. Especially in high-stake scenarios, inherently interpretable models should be preferred \cite{Rudin2019}.

\paragraph{Deriving Knowledge from Explainable Results}
%%%%%%%%%%%%%%%% summary %%%%%%%%%%%%%%%%%
In \autoref{subsec:derive} knowledge is derived from explainability methods which is subsequently integrated into the machine learning pipeline.

%%%%%%%%%%%%%%%% key strengths %%%%%%%%%%%%%%%%
The common way to improve a model is an iterative trial-and-error approach, e.g., feature engineering, in which the knowledge that is gained %from the explanations
is rather implicit.
In contrast, the methods that were reviewed make explicit use of explainability methods to generate insights that are formalized and then used to inform the model in the next iteration.
This means that explainability is a precursor for informed machine learning.
Consequently, this could give rise to an improved methodology where the formalization enables a more explicit way to elaborate on the reasoning for certain design choices. This can help decoupling seemingly arbitrary decision made in a specific context to  generate insight on broader scale.

%%%%%%%%%%%%%%%% future work %%%%%%%%%%%%%%%%
We found that very few methods incorporate the gained knowledge from the explanation in, for example, the model architecture, suggesting an opportunity for future work.
For future research in this domain we refer to recent work \cite{wiegreffe2021teach} which provides data sets for explainability research.

%%%%%%%%%%%%%%%% summary %%%%%%%%%%%%%%%%%
\paragraph{In Summary} The benefits of integrating prior knowledge for explainability can be summarized as the improvement of human understanding of the machine learning model and capturing desired desiderata in the learning process. The greatest challenge lies in the evaluation of explainability which requires further work to assess the effects of different knowledge integration techniques.
A general knowledge source can be used as a mediator to provide the necessary context for a user to interact with a system in a meaningful way. For this reason, it would be highly beneficial for novel explainability methods to provide a designated interface to incorporate context specific knowledge.

\paragraph{Open Research Directions}

We see the following open research directions:
\begin{itemize}
    \item Knowledge formalization: Transforming intuitive knowledge in the form of human feedback to other knowledge representations
    \item Research on the precise effect of knowledge integration on explainability
    \item Make implicit development processes explicit through formalization of applied priors to facilitate development and adaptability of future research
    \item Designated interfaces to incorporate context specific knowledge into explainability methods
\end{itemize}

\section{Conclusion}
\label{sec:conclusion}
To bring machine learning models to the level of being versatile and applicable, explainability is an essential component.
Most \ml approaches are data-constrained and can only provide explanations stemming from the information in the training data.
Hence, we propose to harness prior knowledge such as logical rules or knowledge graphs with the goal to improve explainability.
In this paper, we presented three approaches to integrate prior knowledge into the \ml pipeline and in the explainability component.
The three approaches can be distinguished between the integration method and how prior knowledge is obtained:
Prior knowledge is available independently of data/pipeline and is (1) incorporated in the \ml pipeline or (2) in the explainability method. Prior knowledge can also be (3) derived from model explanations, formalized and then incorporated in the pipeline.
With this we have created a structure that serves for orientation.
Further research is needed to formalize and measure to what extent knowledge integration improves explainability.

%% file: main.bbl
\begin{thebibliography}{86}
\providecommand{\natexlab}[1]{#1}
\providecommand{\url}[1]{\texttt{#1}}
\expandafter\ifx\csname urlstyle\endcsname\relax
  \providecommand{\doi}[1]{doi: #1}\else
  \providecommand{\doi}{doi: \begingroup \urlstyle{rm}\Url}\fi

\bibitem[reg(2016)]{regulation2016regulation}
{Regulation (EU) 2016/679 of the European Parliament and of the Council of 27
  April 2016 on the protection of natural persons with regard to the processing
  of personal data and on the free movement of such data, and repealing
  Directive 95/46/EC (General Data Protection Regulation)}, 2016.

\bibitem[Adadi and Berrada(2018)]{adadi2018peeking}
A.~Adadi and M.~Berrada.
\newblock Peeking inside the black-box: A survey on explainable artificial
  intelligence ({XAI}).
\newblock \emph{IEEE Access}, 6:\penalty0 52138--52160, 2018.

\bibitem[Arrieta et~al.(2020)Arrieta, D{\'\i}az-Rodr{\'\i}guez, Del~Ser,
  Bennetot, Tabik, Barbado, Garc{\'\i}a, Gil-L{\'o}pez, Molina, Benjamins,
  et~al.]{arrieta2020explainable}
A.~B. Arrieta, N.~D{\'\i}az-Rodr{\'\i}guez, J.~Del~Ser, A.~Bennetot, S.~Tabik,
  A.~Barbado, S.~Garc{\'\i}a, S.~Gil-L{\'o}pez, D.~Molina, R.~Benjamins, et~al.
\newblock Explainable artificial intelligence (xai): Concepts, taxonomies,
  opportunities and challenges toward responsible ai.
\newblock \emph{Information Fusion}, 58:\penalty0 82--115, 2020.

\bibitem[Aumann and Shapley(1974)]{Aumann1974}
R.~J. Aumann and L.~S. Shapley.
\newblock \emph{{Values of Non-Atomic Games}}.
\newblock Princeton University Press, 1974.

\bibitem[Balayan et~al.(2020)Balayan, Saleiro, Bel{\'{e}}m, Krippahl, and
  Bizarro]{Balayan2020}
V.~Balayan, P.~Saleiro, C.~Bel{\'{e}}m, L.~Krippahl, and P.~Bizarro.
\newblock {Teaching the Machine to Explain Itself using Domain Knowledge}.
\newblock \emph{arXiv preprint arXiv:2012.01932}, 2020.

\bibitem[Baur et~al.(2020)Baur, Heimerl, Lingenfelser, Wagner, Valstar,
  Schuller, and Andr{\'{e}}]{Baur2020}
T.~Baur, A.~Heimerl, F.~Lingenfelser, J.~Wagner, M.~F. Valstar, B.~Schuller,
  and E.~Andr{\'{e}}.
\newblock {eXplainable Cooperative Machine Learning with NOVA}.
\newblock \emph{KI - K{\"{u}}nstliche Intelligenz}, 34\penalty0 (2):\penalty0
  143--164, 2020.

\bibitem[Bikmukhametov and J{\"a}schke(2020)]{bikmukhametov2020combining}
T.~Bikmukhametov and J.~J{\"a}schke.
\newblock Combining machine learning and process engineering physics towards
  enhanced accuracy and explainability of data-driven models.
\newblock \emph{Computers \& Chemical Engineering}, 138:\penalty0 106834, 2020.

\bibitem[{Binder, Alexander Montavon} et~al.(2016){Binder, Alexander Montavon},
  Lapuschkin, M{\"{u}}ller, and Samek]{BinderAlexanderMontavon2016}
G.~{Binder, Alexander Montavon}, S.~Lapuschkin, K.-R. M{\"{u}}ller, and
  W.~Samek.
\newblock {Layer-Wise Relevance Propagation for Neural Networks with Local
  Renormalization Layers}.
\newblock In \emph{Artificial Neural Networks and Machine Learning -- ICANN
  2016}, Lecture Notes in Computer Science, pages 63--71. Springer
  International Publishing, 2016.

\bibitem[Bouraoui et~al.(2019)Bouraoui, Cornu{\'{e}}jols, Den{\oe}ux,
  Destercke, Dubois, Guillaume, Marques-Silva, Mengin, Prade, Schockaert,
  Serrurier, and Vrain]{Bouraoui2019}
Z.~Bouraoui, A.~Cornu{\'{e}}jols, T.~Den{\oe}ux, S.~Destercke, D.~Dubois,
  R.~Guillaume, J.~Marques-Silva, J.~Mengin, H.~Prade, S.~Schockaert,
  M.~Serrurier, and C.~Vrain.
\newblock {From Shallow to Deep Interactions Between Knowledge Representation,
  Reasoning and Machine Learning}.
\newblock \emph{arXiv preprint arXiv:1912.06612}, dec 2019.

\bibitem[Brundage et~al.(2020)Brundage, Avin, Wang, Belfield, Krueger,
  Hadfield, Khlaaf, Yang, Toner, Fong, et~al.]{brundage2020toward}
M.~Brundage, S.~Avin, J.~Wang, H.~Belfield, G.~Krueger, G.~Hadfield, H.~Khlaaf,
  J.~Yang, H.~Toner, R.~Fong, et~al.
\newblock Toward trustworthy {AI} development: mechanisms for supporting
  verifiable claims.
\newblock \emph{arXiv preprint arXiv:2004.07213}, 2020.

\bibitem[Bu\c{c}inca et~al.(2020)Bu\c{c}inca, Lin, Gajos, and
  Glassman]{buccinca2020}
Z.~Bu\c{c}inca, P.~Lin, K.~Z. Gajos, and E.~L. Glassman.
\newblock Proxy tasks and subjective measures can be misleading in evaluating
  explainable ai systems.
\newblock In \emph{Proceedings of the 25th International Conference on
  Intelligent User Interfaces}, IUI '20, page 454–464, New York, NY, USA,
  2020. Association for Computing Machinery.

\bibitem[Burkart and Huber(2021)]{Burkart2021}
N.~Burkart and M.~F. Huber.
\newblock A survey on the explainability of supervised machine learning.
\newblock \emph{Journal of Artificial Intelligence Research}, 70:\penalty0
  245--317, May 2021.

\bibitem[Camburu et~al.(2018)Camburu, Rockt{\"{a}}schel, Lukasiewicz, and
  Blunsom]{Camburu2018}
O.~M. Camburu, T.~Rockt{\"{a}}schel, T.~Lukasiewicz, and P.~Blunsom.
\newblock {E-SNLI: Natural language inference with natural language
  explanations}.
\newblock \emph{Advances in Neural Information Processing Systems}, \penalty0
  (NeurIPS):\penalty0 9539--9549, 2018.

\bibitem[Chari et~al.(2020)Chari, Gruen, Seneviratne, and
  McGuinness]{chari2020directions}
S.~Chari, D.~Gruen, O.~Seneviratne, and D.~McGuinness.
\newblock Directions for explainable knowledge-enabled systems.
\newblock In \emph{Knowledge Graphs for eXplainable Artificial Intelligence},
  2020.

\bibitem[Chen et~al.(2012)Chen, Zhou, and Prasanna]{Chen2012}
N.~Chen, Q.~Y. Zhou, and V.~K. Prasanna.
\newblock {Understanding web images by object relation network}.
\newblock In \emph{Proceedings of the 21st International Conference on World
  Wide Web}, WWW '12, pages 291--300, New York, NY, USA, 2012. Association for
  Computing Machinery.

\bibitem[Chen et~al.(2020)Chen, Bei, and Rudin]{Chen2020}
Z.~Chen, Y.~Bei, and C.~Rudin.
\newblock {Concept whitening for interpretable image recognition}.
\newblock \emph{Nature Machine Intelligence}, 2\penalty0 (12):\penalty0
  772--782, 2020.

\bibitem[Choi et~al.(2017)Choi, Bahadori, Song, Stewart, and Sun]{Choi2017}
E.~Choi, M.~T. Bahadori, L.~Song, W.~F. Stewart, and J.~Sun.
\newblock {GRAM: Graph-based attention model for healthcare representation
  learning}.
\newblock \emph{Proceedings of the ACM SIGKDD International Conference on
  Knowledge Discovery and Data Mining}, pages 787--795, 2017.

\bibitem[Confalonieri et~al.(2021)Confalonieri, Coba, Wagner, and
  Besold]{confalonieri2021historical}
R.~Confalonieri, L.~Coba, B.~Wagner, and T.~R. Besold.
\newblock A historical perspective of explainable artificial intelligence.
\newblock \emph{WIREs Data Mining and Knowledge Discovery}, 11\penalty0 (1),
  2021.

\bibitem[Dandl et~al.(2020)Dandl, Molnar, Binder, and Bischl]{Dandl2020}
S.~Dandl, C.~Molnar, M.~Binder, and B.~Bischl.
\newblock Multi-objective counterfactual explanations.
\newblock In T.~B{\"a}ck, M.~Preuss, A.~Deutz, H.~Wang, C.~Doerr, M.~Emmerich,
  and H.~Trautmann, editors, \emph{Parallel Problem Solving from Nature -- PPSN
  XVI}, pages 448--469, Cham, 2020. Springer International Publishing.

\bibitem[Donadello et~al.(2017)Donadello, Kessler, and Serafini]{Donadello2017}
I.~Donadello, F.~B. Kessler, and L.~Serafini.
\newblock {Semantic Image Interpretation}.
\newblock \emph{arXiv preprint arXiv:1705.08968v1}, pages 1--14, 2017.

\bibitem[Doran et~al.(2017)Doran, Schulz, and Besold]{doran2017does}
D.~Doran, S.~Schulz, and T.~R. Besold.
\newblock What does explainable {AI} really mean? a new conceptualization of
  perspectives.
\newblock \emph{arXiv preprint arXiv:1710.00794}, 2017.

\bibitem[Erion et~al.(2019)Erion, Janizek, Sturmfels, Lundberg, and
  Lee]{Erion2019}
G.~Erion, J.~D. Janizek, P.~Sturmfels, S.~Lundberg, and S.-I. Lee.
\newblock {Improving performance of deep learning models with axiomatic
  attribution priors and expected gradients}.
\newblock \emph{arXiv preprint arXiv:1906.10670}, 2019.

\bibitem[Frosst and Hinton(2017)]{Frosst2017}
N.~Frosst and G.~E. Hinton.
\newblock {Distilling a Neural Network Into a Soft Decision Tree}.
\newblock \emph{arXiv preprint arXiv:1711.09784}, 2017.

\bibitem[Hase and Bansal(2020)]{hase2020-evaluating}
P.~Hase and M.~Bansal.
\newblock Evaluating explainable {AI}: Which algorithmic explanations help
  users predict model behavior?
\newblock In \emph{Proceedings of the 58th Annual Meeting of the Association
  for Computational Linguistics}, pages 5540--5552, Online, July 2020.
  Association for Computational Linguistics.

\bibitem[Jiang et~al.(2019)Jiang, Hewner, and Chandola]{Jiang2019}
J.~Jiang, S.~Hewner, and V.~Chandola.
\newblock Tree-based regularization for interpretable readmission prediction.
\newblock In \emph{Proceedings of the {AAAI} 2019 Spring Symposium on Combining
  Machine Learning with Knowledge Engineering {AAAI-MAKE}}, volume 2350 of
  \emph{{CEUR} Workshop Proceedings}. CEUR-WS.org, 2019.

\bibitem[Karpatne et~al.(2017)Karpatne, Atluri, Faghmous, Steinbach, Banerjee,
  Ganguly, Shekhar, Samatova, and Kumar]{karpatne2017theory}
A.~Karpatne, G.~Atluri, J.~H. Faghmous, M.~Steinbach, A.~Banerjee, A.~Ganguly,
  S.~Shekhar, N.~Samatova, and V.~Kumar.
\newblock Theory-guided data science: A new paradigm for scientific discovery
  from data.
\newblock \emph{IEEE Transactions on Knowledge and Data Engineering},
  29\penalty0 (10):\penalty0 2318--2331, 2017.

\bibitem[Kieseberg et~al.(2015)Kieseberg, Schantl, Fr{\"{u}}hwirt, Weippl, and
  Holzinger]{10.1007/978-3-319-23344-4_36}
P.~Kieseberg, J.~Schantl, P.~Fr{\"{u}}hwirt, E.~Weippl, and A.~Holzinger.
\newblock {Witnesses for the Doctor in the Loop}.
\newblock In \emph{Brain Informatics and Health}, pages 369--378, Cham, 2015.
  Springer International Publishing.

\bibitem[Kim et~al.(2018)Kim, Wattenberg, Gilmer, Cai, Wexler, Viegas, and
  Sayres]{Kim2018a}
B.~Kim, M.~Wattenberg, J.~Gilmer, C.~Cai, J.~Wexler, F.~Viegas, and R.~Sayres.
\newblock {Interpretability beyond feature attribution: Quantitative Testing
  with Concept Activation Vectors (TCAV)}.
\newblock In \emph{35th International Conference on Machine Learning, ICML
  2018}, volume~6 of \emph{Proceedings of Machine Learning Research}, pages
  4186--4195. PMLR, 2018.

\bibitem[Krause et~al.(2016)Krause, Perer, and Ng]{Krause2016}
J.~Krause, A.~Perer, and K.~Ng.
\newblock Interacting with predictions: Visual inspection of black-box machine
  learning models.
\newblock In \emph{Conference on Human Factors in Computing Systems -
  Proceedings}, pages 5686--5697. Association for Computing Machinery, 2016.

\bibitem[Krishnan(2020)]{Krishnan2020a}
M.~Krishnan.
\newblock {Against Interpretability: a Critical Examination of the
  Interpretability Problem in Machine Learning}.
\newblock \emph{Philosophy and Technology}, 33\penalty0 (3):\penalty0 487--502,
  2020.

\bibitem[Kulesza et~al.(2015)Kulesza, Burnett, Wong, and Stumpf]{Kulesza2015}
T.~Kulesza, M.~Burnett, W.~K. Wong, and S.~Stumpf.
\newblock {Principles of Explanatory Debugging to personalize interactive
  machine learning}.
\newblock In \emph{International Conference on Intelligent User Interfaces,
  Proceedings IUI}, pages 126--137. Association for Computing Machinery, 2015.

\bibitem[Lakkaraju et~al.(2019)Lakkaraju, Kamar, Caruana, and
  Leskovec]{Lakkaraju2019}
H.~Lakkaraju, E.~Kamar, R.~Caruana, and J.~Leskovec.
\newblock Faithful and customizable explanations of black box models.
\newblock In \emph{Proceedings of the 2019 AAAI/ACM Conference on AI, Ethics,
  and Society}, AIES '19, page 131–138, New York, NY, USA, 2019. Association
  for Computing Machinery.

\bibitem[Lapuschkin et~al.(2019)Lapuschkin, Wäldchen, Binder, Montavon, Samek,
  and Müller]{lapuschkin_unmasking_2019}
S.~Lapuschkin, S.~Wäldchen, A.~Binder, G.~Montavon, W.~Samek, and K.-R.
  Müller.
\newblock Unmasking {Clever} {Hans} predictors and assessing what machines
  really learn.
\newblock \emph{Nature Communications}, 10\penalty0 (1):\penalty0 1096, Dec.
  2019.

\bibitem[Lauer and Bloch(2008)]{lauer2008incorporating}
F.~Lauer and G.~Bloch.
\newblock Incorporating prior knowledge in support vector machines for
  classification: A review.
\newblock \emph{Neurocomputing}, 71\penalty0 (7):\penalty0 1578--1594, 2008.

\bibitem[Li et~al.(2020)Li, Cao, Shi, Bai, Gao, Qiu, Wang, Gao, Zhang, Xue, and
  Chen]{Li2020}
X.-H. Li, C.~C. Cao, Y.~Shi, W.~Bai, H.~Gao, L.~Qiu, C.~Wang, Y.~Gao, S.~Zhang,
  X.~Xue, and L.~Chen.
\newblock {A Survey of Data-driven and Knowledge-aware eXplainable AI}.
\newblock \emph{IEEE Transactions on Knowledge and Data Engineering}, 2020.

\bibitem[Lipton(2018)]{Lipton2018a}
Z.~C. Lipton.
\newblock The mythos of model interpretability: In machine learning, the
  concept of interpretability is both important and slippery.
\newblock \emph{Queue}, 16\penalty0 (3):\penalty0 31–57, June 2018.

\bibitem[Liu and Avci(2019)]{Liu2019a}
F.~Liu and B.~Avci.
\newblock Incorporating priors with feature attribution on text classification.
\newblock In \emph{Proceedings of the 57th Annual Meeting of the Association
  for Computational Linguistics}, pages 6274--6283, Florence, Italy, July 2019.
  Association for Computational Linguistics.

\bibitem[Liu et~al.(2017)Liu, Wang, Liu, and Zhu]{liu2017better}
S.~Liu, X.~Wang, M.~Liu, and J.~Zhu.
\newblock {Towards Better Analysis of Machine Learning Models: A Visual
  Analytics Perspective}.
\newblock \emph{Visual Informatics}, 1\penalty0 (1):\penalty0 48--56, 2017.

\bibitem[Liu et~al.(2019)Liu, Niu, Wu, and Wang]{Liu2020}
Z.~Liu, Z.-Y. Niu, H.~Wu, and H.~Wang.
\newblock Knowledge aware conversation generation with explainable reasoning
  over augmented graphs.
\newblock In \emph{Proceedings of the 2019 Conference on Empirical Methods in
  Natural Language Processing and the 9th International Joint Conference on
  Natural Language Processing (EMNLP-IJCNLP)}, pages 1782--1792, Hong Kong,
  China, Nov. 2019. Association for Computational Linguistics.

\bibitem[Lundberg and Lee(2017)]{Lundberg2017}
S.~M. Lundberg and S.-I. Lee.
\newblock {A unified approach to interpreting model predictions}.
\newblock In \emph{Proceedings of the 31st International Conference on Neural
  Information Processing Systems}, number Section 2, pages 1--10, 2017.

\bibitem[Ma et~al.(2018)Ma, You, Xiao, Chitta, Zhou, and Gao]{Ma2018c}
F.~Ma, Q.~You, H.~Xiao, R.~Chitta, J.~Zhou, and J.~Gao.
\newblock Kame: Knowledge-based attention model for diagnosis prediction in
  healthcare.
\newblock In \emph{Proceedings of the 27th ACM International Conference on
  Information and Knowledge Management}, CIKM '18, page 743–752, New York,
  NY, USA, 2018. Association for Computing Machinery.

\bibitem[Ma and Zhang(2019)]{Ma2019a}
T.~Ma and A.~Zhang.
\newblock {Incorporating Biological Knowledge with Factor Graph Neural Network
  for Interpretable Deep Learning}.
\newblock \emph{arXiv preprint arXiv:1906.00537}, 2019.

\bibitem[Ma et~al.(2019)Ma, Jin, Zhang, Wang, Cao, Liu, Ma, and Ren]{Ma2019}
W.~Ma, W.~Jin, M.~Zhang, C.~Wang, Y.~Cao, Y.~Liu, S.~Ma, and X.~Ren.
\newblock {Jointly learning explainable rules for recommendation with knowledge
  graph}.
\newblock \emph{The Web Conference 2019 - Proceedings of the World Wide Web
  Conference, WWW 2019}, pages 1210--1221, 2019.

\bibitem[Mahajan et~al.(2019)Mahajan, Tan, and Sharma]{Mahajan2019}
D.~Mahajan, C.~Tan, and A.~Sharma.
\newblock {Preserving Causal Constraints in Counterfactual Explanations for
  Machine Learning Classifiers}.
\newblock \emph{arXiv preprint arXiv:1912.03277}, 2019.

\bibitem[Mahendran and Vedaldi(2016)]{Mahendran2015}
A.~Mahendran and A.~Vedaldi.
\newblock {Visualizing Deep Convolutional Neural Networks Using Natural
  Pre-Images}.
\newblock \emph{International Journal of Computer Vision}, 120\penalty0
  (3):\penalty0 233--255, dec 2016.
\newblock \doi{10.1007/s11263-016-0911-8}.

\bibitem[Miller(2019)]{Miller2019}
T.~Miller.
\newblock Explanation in artificial intelligence: Insights from the social
  sciences.
\newblock \emph{Artificial Intelligence}, 267:\penalty0 1--38, 2019.

\bibitem[Molnar et~al.(2020)Molnar, K{\"{o}}nig, Herbinger, Freiesleben, Dandl,
  Scholbeck, Casalicchio, Grosse-Wentrup, and Bischl]{Molnar2020a}
C.~Molnar, G.~K{\"{o}}nig, J.~Herbinger, T.~Freiesleben, S.~Dandl, C.~A.
  Scholbeck, G.~Casalicchio, M.~Grosse-Wentrup, and B.~Bischl.
\newblock {Pitfalls to Avoid when Interpreting Machine Learning Models}.
\newblock \emph{arXiv preprint arXiv:2007.04131}, 2020.

\bibitem[Mothilal et~al.(2020)Mothilal, Sharma, and Tan]{Mothilal2020}
R.~K. Mothilal, A.~Sharma, and C.~Tan.
\newblock {Explaining machine learning classifiers through diverse
  counterfactual explanations}.
\newblock In \emph{FAT* 2020 - Proceedings of the 2020 Conference on Fairness,
  Accountability, and Transparency}, pages 607--617. Association for Computing
  Machinery, Inc, 2020.

\bibitem[Muggleton and {de Raedt}(1994)]{muggleton1994}
S.~Muggleton and L.~{de Raedt}.
\newblock Inductive logic programming: Theory and methods.
\newblock \emph{The Journal of Logic Programming}, 19-20:\penalty0 629--679,
  1994.
\newblock ISSN 0743-1066.
\newblock Special Issue: Ten Years of Logic Programming.

\bibitem[Muggleton et~al.(2018)Muggleton, Schmid, Zeller, Tamaddoni-Nezhad, and
  Besold]{muggleton2018}
S.~H. Muggleton, U.~Schmid, C.~Zeller, A.~Tamaddoni-Nezhad, and T.~Besold.
\newblock Ultra-strong machine learning: comprehensibility of programs learned
  with {ILP}.
\newblock \emph{Machine Learning}, 107\penalty0 (7):\penalty0 1119--1140, 2018.

\bibitem[Norkute(2021)]{norkute2021}
M.~Norkute.
\newblock {AI} explainability: Why one explanation cannot fit all.
\newblock In \emph{ACM CHI Workshop on Operationalizing Human-Centered
  Perspectives in Explainable AI (HCXAI)}, 2021.

\bibitem[Paiva et~al.(2015)Paiva, Schwartz, Pedrini, and Minghim]{Paiva2015}
J.~G.~S. Paiva, W.~R. Schwartz, H.~Pedrini, and R.~Minghim.
\newblock {An approach to supporting incremental visual data classification}.
\newblock \emph{IEEE Transactions on Visualization and Computer Graphics},
  21\penalty0 (1):\penalty0 4--17, 2015.

\bibitem[Pommellet and L{\'{e}}cu{\'{e}}(2019)]{Pommellet2019}
T.~Pommellet and F.~L{\'{e}}cu{\'{e}}.
\newblock {Feeding machine learning with knowledge graphs for explainable
  object detection}.
\newblock \emph{CEUR Workshop Proceedings}, 2456:\penalty0 277--280, 2019.

\bibitem[Rabold et~al.(2019)Rabold, Deininger, Siebers, and
  Schmid]{rabold2019enriching}
J.~Rabold, H.~Deininger, M.~Siebers, and U.~Schmid.
\newblock Enriching visual with verbal explanations for relational
  concepts--combining lime with aleph.
\newblock In \emph{Proceedings of the Joint European Conference on Machine
  Learning and Knowledge Discovery in Databases (ECML-PKDD)}, pages 180--192.
  Springer, 2019.

\bibitem[Raissi et~al.(2019)Raissi, Perdikaris, and
  Karniadakis]{raissi2019physics}
M.~Raissi, P.~Perdikaris, and G.~E. Karniadakis.
\newblock Physics-informed neural networks: A deep learning framework for
  solving forward and inverse problems involving nonlinear partial differential
  equations.
\newblock \emph{Journal of Computational Physics}, 378:\penalty0 686--707,
  2019.

\bibitem[Ras et~al.(2018)Ras, van Gerven, and Haselager]{ras2018explanation}
G.~Ras, M.~van Gerven, and P.~Haselager.
\newblock \emph{Explanation Methods in Deep Learning: Users, Values, Concerns
  and Challenges}, pages 19--36.
\newblock Springer International Publishing, Cham, 2018.

\bibitem[Ribeiro et~al.(2016)Ribeiro, Singh, and Guestrin]{Ribeiro2016}
M.~T. Ribeiro, S.~Singh, and C.~Guestrin.
\newblock {"Why Should I Trust You?": Explaining the Predictions of Any
  Classifier}.
\newblock In \emph{Proceedings of the 22nd ACM SIGKDD International Conference
  on Knowledge Discovery and Data Mining}, KDD '16, pages 1135--1144, San
  Francisco, California, USA, 2016. Association for Computing Machinery.

\bibitem[Rieger et~al.(2020)Rieger, Singh, Murdoch, and Yu]{rieger20a}
L.~Rieger, C.~Singh, W.~Murdoch, and B.~Yu.
\newblock Interpretations are useful: Penalizing explanations to align neural
  networks with prior knowledge.
\newblock In H.~D. III and A.~Singh, editors, \emph{Proceedings of the 37th
  International Conference on Machine Learning}, volume 119 of
  \emph{Proceedings of Machine Learning Research}, pages 8116--8126. PMLR,
  13--18 Jul 2020.

\bibitem[Roscher et~al.(2020)Roscher, Bohn, Duarte, and Garcke]{Roscher2019}
R.~Roscher, B.~Bohn, M.~F. Duarte, and J.~Garcke.
\newblock {Explainable Machine Learning for Scientific Insights and
  Discoveries}.
\newblock \emph{IEEE Access}, 8:\penalty0 42200--42216, 2020.

\bibitem[Ross et~al.(2017)Ross, Hughes, and Doshi-Velez]{Ross2017}
A.~S. Ross, M.~C. Hughes, and F.~Doshi-Velez.
\newblock {Right for the right reasons: Training differentiable models by
  constraining their explanations}.
\newblock In \emph{IJCAI International Joint Conference on Artificial
  Intelligence}, 2017.

\bibitem[Rudin(2019)]{Rudin2019}
C.~Rudin.
\newblock {Stop explaining black box machine learning models for high stakes
  decisions and use interpretable models instead}.
\newblock \emph{Nature Machine Intelligence}, 1\penalty0 (5):\penalty0
  206--215, 2019.

\bibitem[Rybakov et~al.(2020)Rybakov, Lotfollahi, Theis, and Wolf]{Rybakov2020}
S.~Rybakov, M.~Lotfollahi, F.~J. Theis, and F.~A. Wolf.
\newblock {Learning interpretable latent autoencoder representations with
  annotations of feature sets}.
\newblock \emph{bioRxiv}, 2020.

\bibitem[Samek et~al.(2017)Samek, Wiegand, and M{\"{u}}ller]{Samek2017XAI}
W.~Samek, T.~Wiegand, and K.-R. M{\"{u}}ller.
\newblock {Explainable Artificial Intelligence: Understanding, Visualizing and
  Interpreting Deep Learning Models}.
\newblock \emph{arXiv preprint arXiv:1708.08296}, 2017.

\bibitem[Schneider and Handali(2019)]{schneider2019personalized}
J.~Schneider and J.~Handali.
\newblock {Personalized explanation in machine learning: A conceptualization}.
\newblock \emph{arXiv preprint arXiv:1901.00770}, 2019.

\bibitem[Schramowski et~al.(2020)Schramowski, Stammer, Teso, Brugger, Herbert,
  Shao, Luigs, Mahlein, and Kersting]{Schramowski2020a}
P.~Schramowski, W.~Stammer, S.~Teso, A.~Brugger, F.~Herbert, X.~Shao, H.~G.
  Luigs, A.~K. Mahlein, and K.~Kersting.
\newblock {Making deep neural networks right for the right scientific reasons
  by interacting with their explanations}.
\newblock \emph{Nature Machine Intelligence}, 2\penalty0 (8):\penalty0
  476--486, 2020.

\bibitem[Selvaraju et~al.(2019)Selvaraju, Lee, Shen, Jin, Ghosh, Heck, Batra,
  and Parikh]{Selvaraju2019}
R.~R. Selvaraju, S.~Lee, Y.~Shen, H.~Jin, S.~Ghosh, L.~Heck, D.~Batra, and
  D.~Parikh.
\newblock {Taking a HINT: Leveraging explanations to make vision and language
  models more grounded}.
\newblock In \emph{Proceedings of the IEEE International Conference on Computer
  Vision}, pages 2591--2600, 2019.

\bibitem[Shams et~al.(2021)Shams, Dimanov, Kola, Simidjievski, Terre, Scherer,
  Matjasec, Abraham, Lio, and Jamnik]{shams2021rem}
Z.~Shams, B.~Dimanov, S.~Kola, N.~Simidjievski, H.~A. Terre, P.~Scherer,
  U.~Matjasec, J.~Abraham, P.~Lio, and M.~Jamnik.
\newblock {REM}: An integrative rule extraction methodology for explainable
  data analysis in healthcare.
\newblock \emph{medRxiv}, 2021.

\bibitem[Shapley(1953)]{Shapley1953}
L.~S. Shapley.
\newblock {A value for n-person games}.
\newblock \emph{Contributions to the Theory of Games}, 2\penalty0
  (28):\penalty0 307--317, 1953.

\bibitem[Shivaswamy and Joachims(2015)]{Shivaswamy2015}
P.~Shivaswamy and T.~Joachims.
\newblock {Coactive Learning}.
\newblock \emph{Journal of Artificial Intelligence Research}, 53:\penalty0
  1--40, 2015.

\bibitem[Sokol and Flach(2018)]{Sokol2017}
K.~Sokol and P.~Flach.
\newblock Glass-box: Explaining ai decisions with counterfactual statements
  through conversation with a voice-enabled virtual assistant.
\newblock In \emph{Proceedings of the Twenty-Seventh International Joint
  Conference on Artificial Intelligence, {IJCAI-18}}, pages 5868--5870.
  International Joint Conferences on Artificial Intelligence Organization, 7
  2018.

\bibitem[Sokol and Flach(2019)]{Sokol2019}
K.~Sokol and P.~Flach.
\newblock {Explainability Fact Sheets: A Framework for Systematic Assessment of
  Explainable Approaches}.
\newblock \emph{FAT* 2020 - Proceedings of the 2020 Conference on Fairness,
  Accountability, and Transparency}, pages 56--67, 2019.

\bibitem[Sokol and Flach(2020)]{Sokol_2020}
K.~Sokol and P.~Flach.
\newblock {One Explanation Does Not Fit All}.
\newblock \emph{KI - K{\"{u}}nstliche Intelligenz}, 34\penalty0 (2):\penalty0
  235--250, 2020.

\bibitem[Stammer et~al.(2020)Stammer, Schramowski, and Kersting]{Stammer2020}
W.~Stammer, P.~Schramowski, and K.~Kersting.
\newblock {Right for the right concept: Revising neuro-symbolic concepts by
  interacting with their explanations}.
\newblock \emph{arXiv preprint arXiv:2011.12854}, 2020.

\bibitem[Stepin et~al.(2021)Stepin, Alonso, Catala, and
  Pereira-Fari\~na]{stepin2021cf}
I.~Stepin, J.~M. Alonso, A.~Catala, and M.~Pereira-Fari\~na.
\newblock A survey of contrastive and counterfactual explanation generation
  methods for explainable artificial intelligence.
\newblock \emph{IEEE Access}, 9:\penalty0 11974--12001, 2021.

\bibitem[Sundararajan et~al.(2017)Sundararajan, Taly, and
  Yan]{Sundararajan2017}
M.~Sundararajan, A.~Taly, and Q.~Yan.
\newblock {Axiomatic attribution for deep networks}.
\newblock In \emph{Proceedings of the 34th International Conference on Machine
  Learning, ICML 2017}, volume~7 of \emph{ICML'17}, pages 5109--5118, Sydney,
  NSW, Australia, 2017. JMLR.org.

\bibitem[Tan et~al.(2018)Tan, Caruana, Hooker, Koch, and Gordo]{Tan2018}
S.~Tan, R.~Caruana, G.~Hooker, P.~Koch, and A.~Gordo.
\newblock {Learning Global Additive Explanations for Neural Nets Using Model
  Distillation}.
\newblock \emph{arXiv:1801.08640}, 2018.

\bibitem[Teso and Kersting(2019)]{Teso2019}
S.~Teso and K.~Kersting.
\newblock {Explanatory Interactive Machine Learning}.
\newblock In \emph{AIES 2019 - Proceedings of the 2019 AAAI/ACM Conference on
  AI, Ethics, and Society}, AIES '19, pages 239--245, New York, NY, USA, 2019.
  Association for Computing Machinery.

\bibitem[{Von Rueden} et~al.(2021){Von Rueden}, Mayer, Beckh, Georgiev,
  Giesselbach, Heese, Kirsch, Pfrommer, Pick, Ramamurthy, Walczak, Garcke,
  Bauckhage, and Schuecker]{VonRueden}
L.~{Von Rueden}, S.~Mayer, K.~Beckh, B.~Georgiev, S.~Giesselbach, R.~Heese,
  B.~Kirsch, J.~Pfrommer, A.~Pick, R.~Ramamurthy, M.~Walczak, J.~Garcke,
  C.~Bauckhage, and J.~Schuecker.
\newblock {Informed machine learning - a taxonomy and survey of integrating
  knowledge into learning systems}.
\newblock \emph{IEEE Transactions on Knowledge and Data Engineering}, 2021.

\bibitem[Wachter et~al.(2017)Wachter, Mittelstadt, and Russell]{Wachter2017}
S.~Wachter, B.~Mittelstadt, and C.~Russell.
\newblock {Counterfactual Explanations Without Opening the Black Box: Automated
  Decisions and the GDPR}.
\newblock \emph{Harvard Journal of Law {\&} Technology}, 31\penalty0 (2), 2017.

\bibitem[Wang et~al.(2019)Wang, Wang, Xu, He, Cao, and Chua]{Wang2019d}
X.~Wang, D.~Wang, C.~Xu, X.~He, Y.~Cao, and T.-S. Chua.
\newblock {Explainable reasoning over knowledge graphs for recommendation}.
\newblock \emph{Proceedings of the AAAI Conference on Artificial Intelligence},
  33\penalty0 (01):\penalty0 5329--5336, 2019.

\bibitem[Wei et~al.(2015)Wei, Zhou, Torrabla, and Freeman]{WeiZTF15}
D.~Wei, B.~Zhou, A.~Torrabla, and W.~Freeman.
\newblock Understanding intra-class knowledge inside {CNN}.
\newblock \emph{arXiv preprint arXiv:1507.02379}, 2015.

\bibitem[Wexler et~al.(2020)Wexler, Pushkarna, Bolukbasi, Wattenberg,
  Vi{\'{e}}gas, and Wilson]{8807255}
J.~Wexler, M.~Pushkarna, T.~Bolukbasi, M.~Wattenberg, F.~Vi{\'{e}}gas, and
  J.~Wilson.
\newblock {The What-If Tool: Interactive Probing of Machine Learning Models}.
\newblock \emph{IEEE Transactions on Visualization and Computer Graphics},
  26\penalty0 (1):\penalty0 56--65, 2020.

\bibitem[Wiegreffe and Marasovi{\'c}(2021)]{wiegreffe2021teach}
S.~Wiegreffe and A.~Marasovi{\'c}.
\newblock Teach me to explain: A review of datasets for explainable nlp.
\newblock \emph{arXiv preprint arXiv:2102.12060}, 2021.

\bibitem[Yan et~al.(2019)Yan, Peng, Sandfort, Bagheri, Lu, and
  Summers]{Yan2019}
K.~Yan, Y.~Peng, V.~Sandfort, M.~Bagheri, Z.~Lu, and R.~M. Summers.
\newblock {Holistic and comprehensive annotation of clinically significant
  findings on diverse {CT} images: Learning from radiology reports and label
  ontology}.
\newblock \emph{Proceedings of the IEEE Computer Society Conference on Computer
  Vision and Pattern Recognition}, 2019-June:\penalty0 8515--8524, 2019.

\bibitem[Yosinski et~al.(2015)Yosinski, Clune, Nguyen, Fuchs, and
  Lipson]{Yosinski2015}
J.~Yosinski, J.~Clune, A.~Nguyen, T.~Fuchs, and H.~Lipson.
\newblock {Understanding Neural Networks Through Deep Visualization}.
\newblock \emph{arXiv preprint arXiv:1506.06579}, 2015.

\bibitem[Zhang et~al.(2017)Zhang, Cao, Zhang, Redmonds, Wu, and Zhu]{Zhang2017}
Q.~Zhang, R.~Cao, S.~Zhang, M.~Redmonds, Y.~N. Wu, and S.-C. Zhu.
\newblock {Interactively Transferring CNN Patterns for Part Localization}.
\newblock \emph{arXiv preprint arXiv:1708.01783}, 2017.

\end{thebibliography}
